\documentclass{article}

\usepackage{PRIMEarxiv}

\usepackage[utf8]{inputenc} 
\usepackage[T1]{fontenc}    
\usepackage{hyperref}       
\usepackage{url}            
\usepackage{booktabs}       
\usepackage{amsfonts}       
\usepackage{nicefrac}       
\usepackage{microtype}      
\usepackage{lipsum}
\usepackage{fancyhdr}       
\usepackage{graphicx}       
\graphicspath{{media/}}     
\usepackage{natbib}
\usepackage{rotating}

\usepackage[dvipsnames]{xcolor}
\usepackage{enumitem}
\usepackage{subcaption}
\usepackage{booktabs}
\usepackage{wrapfig}
\usepackage{multirow}
\usepackage{array}
\usepackage{doi}

 \title{SeFNet: Bridging Tabular Datasets with Semantic Feature Nets}

\author{%
  Katarzyna Woźnica \quad Piotr Wilczyński   \quad Przemysław Biecek \\
    Faculty of~Mathematics and Information Sciences\\
  Warsaw University of~Technology\\
  Warsaw \\
  \texttt{\{katarzyna.woznica.dokt, piotr.wilczynski.stud, przemyslaw.biecek\}@pw.edu.pl} \\
}

\begin{document}
\maketitle

\begin{abstract}

Machine learning applications cover a~wide range of~predictive tasks in which tabular datasets play a~significant role. However, although they often address similar problems, tabular datasets are typically treated as standalone tasks. The~possibilities of~using previously solved problems are limited due to the~lack of~structured contextual information about their features and the~lack of~understanding of~the~relations between them.
To overcome this limitation, we propose a~new approach called Semantic Feature Net (SeFNet), capturing the~semantic meaning of~the~analyzed tabular features.
By leveraging existing ontologies and domain knowledge, SeFNet opens up new opportunities for sharing insights between diverse predictive tasks. One such opportunity is the~Dataset Ontology-based Semantic Similarity (DOSS) measure, which quantifies the~similarity between datasets using relations across their features.
In this paper, we present an example of~SeFNet prepared for a~collection of~predictive tasks in healthcare, with the~features' relations derived from the~SNOMED-CT ontology. The~proposed SeFNet framework and the~accompanying DOSS measure address the~issue of~limited contextual information in tabular datasets. By incorporating domain knowledge and establishing semantic relations between features, we enhance the~potential for meta-learning and enable valuable insights to be shared across different predictive tasks.

\end{abstract}

\keywords{tabular datasets \and semantic similarity \and dataset repositories \and meta-learning \and ontology \and healthcare}

\section{Introduction}
\label{sec:intro}

Tabular datasets play a~significant role in machine learning (ML) applications since they are the~most common data type~\citep{mckinsey}. 
Their prevalence results in a great diversity of~numbers and types of~features. 
Each dataset can include a~different set of~variables such as age, gender, income, or education. Thus, we generally refer to a~heterogeneous feature space~\citep{iwata2020meta} when considering a~broad set of~tabular data. Because of~the~features' heterogeneity, most tabular datasets remain unrelated, lacking established relationships to assess their similarity in meaning. This lack of~structured semantic information about datasets is a relevant constraint in the development of~meta-learning methods for tabular tasks.

To fill these gaps, this research paper proposes  \textbf{Semantic Feature Net (SeFNet)} to set up semantic relations between disparate datasets. 
In SeFNet, variables create the net of relations based on semantic information extracted from ontology. 
This resource of~related features and datasets may hold significant potential. As of~today, it can assist machine learning specialists in collaborating with domain experts, facilitate the~exploration of~ similar experiments, and leverage prior insights about various stages of~the~data analysis process, such as feature selection, data imputation, or model optimization. In the~long term, SeFNet can serve as a~resource to enhance meta-learning methods ~\citep{vanschoren2019meta} that automatically extract information from machine learning experiments~\citep{feurer2015initializing,Wistuba2016SequentialTuning,hutter2019automated}.  In particular, it can also be an important contribution to the~development of~methods targeting the~heterogeneous feature space~\citep{iwata2020meta,iwata2022sharing}.

SeFNet is primarily designed for domain-centric applications as it evaluates the~semantic similarity of~variables given the~domain knowledge encoded in ontologies. One notable advantage of~the~proposed framework is its versatility, as it can be applied in any domain with existing ontology. To demonstrate this tool, we prepared a~prototype for the healthcare domain.  That~field is an ideal starting point for applying SeFNet. Firstly, medicine requires a~holistic perspective when addressing prediction problems, considering the~complex interplay between various variables. Additionally, medical datasets often exhibit unique characteristics, such as small sample sizes resulting from rare diseases, emphasizing the~need for innovative approaches to extracting meaningful insights~\citep{alaa2018autoprognosis}.  Moreover, the~medical domain offers an extensive collection of~datasets, providing a~rich source of~information for meta-learning analyses. Finally, medical ontologies and knowledge bases can be further developed more actively due to the new applications provided by SeFNet.

\begin{figure}
    \centering
    \includegraphics[width=0.7\textwidth]{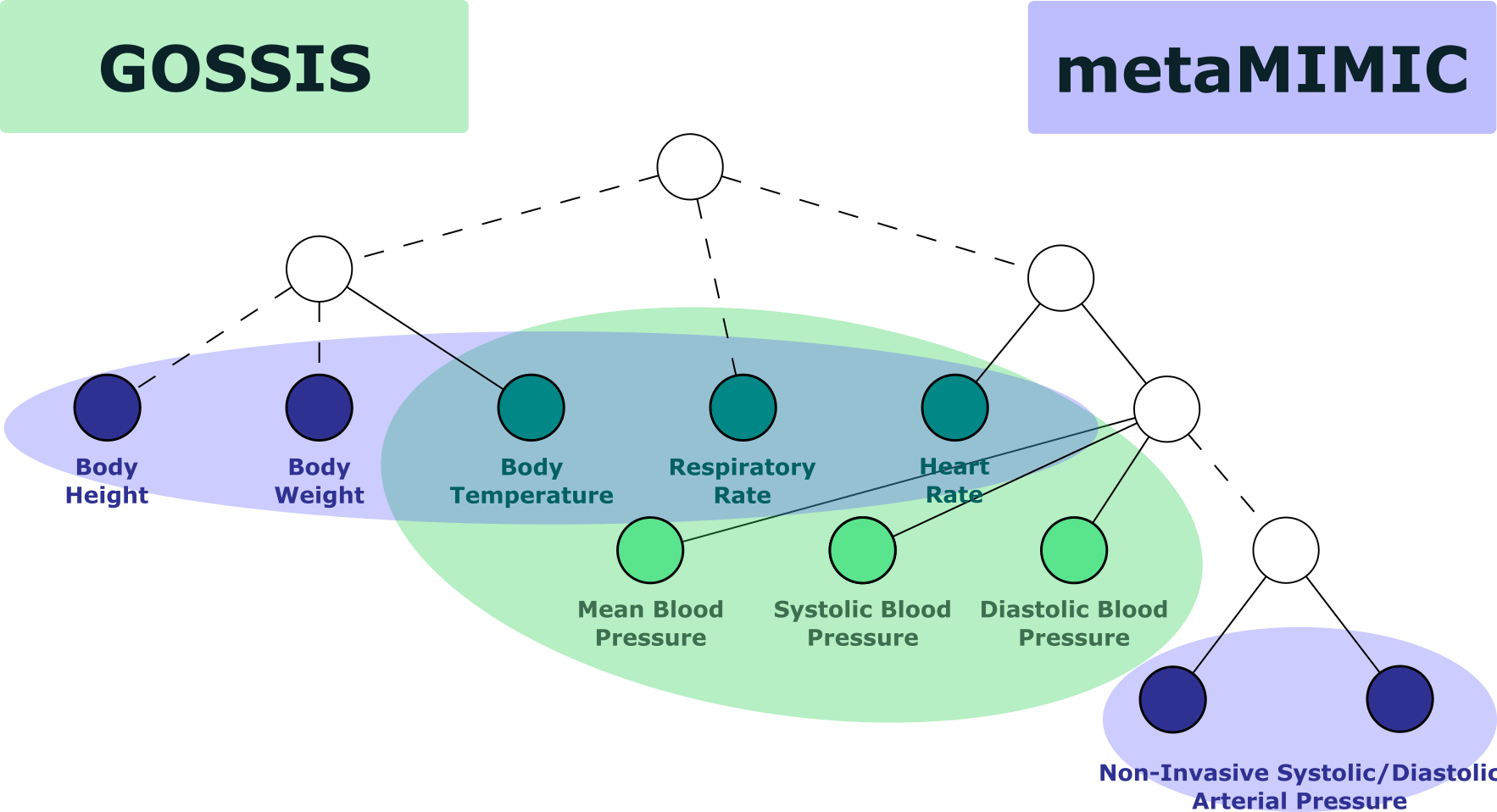}
    \caption{SeFNet for the~two tabular datasets: \textit{metaMIMIC}~\citep{woznica2022consolidated} and \textit{GOSSIS}~\citep{raffa2022global}. This resource encodes the~structure of~the~relations between their features.  Features are mapped on terms from the~SNOMED-CT ontology~\citep{wang2002snomed}.  The~diagram shows the~tree structure of~these features relations. The~blue nodes are features found in the~\textit{metaMIMIC} dataset, and the~green ones are in the~\textit{GOSSIS} dataset. White nodes denote the~common ancestors of~the~presented concepts. A~solid line means that the~lower node is a~direct child of~the~upper node, and a~dashed line means that there are other nodes on the~path between them.}
    \label{fig:my_label}
\end{figure}

\textbf{Contributions.} In this work, (1) we introduce a~Semantic Feature Net (SeFNet) approach that enables the~semantic structuring of~features found in tabular datasets.  Building and applying SeFNet could become a~relevant practice in data collection and curation since it can enable sharing of information about features across diverse tasks and potentially improve meta-learning methods.
The SeFNet approach is highly versatile and can be applied to any ontology, making it widely applicable across various domains. (2) We  created and shared a~comprehensive repository focused specifically on healthcare datasets used in machine learning. The~features within these datasets have been carefully structured and published in SeFNet, resulting in a~collection of~216 distinct features derived from 16 different datasets. This repository can serve as a~valuable resource for researchers and practitioners working in the~healthcare field. (3) We propose a~method for quantifying the~semantic similarity between two datasets by utilizing the similarity of features and employing the~Dataset Ontology-based Semantic Similarity (DOSS) aggregation. The~DOSS representation is a~novel approach that can incorporate semantic meaning into meta-learning methods.

\section{Background and Related work}
\label{sec:rel_work}

\subsection{Datasets used in meta-learning}
\label{sec:rel_data}

Meta-learning is a~broad field focused on the extraction and reuse of~information from machine learning experiments. As resources of~experiments, researchers utilize open repositories containing a wide spectrum of~tasks. Consequently, these repositories contain extensive collections of~tabular datasets, each representing a~separate source of~information.  Various methods have been introduced to establish connections between these datasets, but they lack specificity with regard to domain characteristics. Medical datasets that could be useful in domain-centric approaches are becoming more abundant, yet these datasets remain  standalone.

\textbf{Domain-agnostic collection of~tabular data.}
One of~the~first repositories of~datasets used in benchmarks for tabular data is the~UCI repository ~\citep{Dua:2019}. 
Its origins date back to 1982, and this repository became the~foundation for the~next one, currently the~most widely used  repository OpenML~\citep{OpenML2013}. Since any user can upload their dataset there, OpenML is a~diverse and very broad collection of~datasets. On the~one hand, it is a~good reflection of~the~diversity of~data in various fields, but on the~other hand, the~data is of~very different quality. For this reason, benchmarks using a~fixed subset of~tasks have been created --- OpenML100 and OpenML-CC18~\citep{bischl2017openml100}. A~separate benchmark has been determined for AutoML research~\citep{ambl}. A~parallel initiative was the~creation of~the~Kaggle platform\footnote{\url{https://www.kaggle.com/}}, where anyone can create a~challenge with a~specific prediction problem. Its primary purpose is to allow different teams to compete in preparation of~the~most accurate solution to a~problem, but in addition, Kaggle also became a~vast data repository. Many people use mentioned repositories to obtain data for benchmarking new machine learning methods, but they are treated mostly as standalone tasks, and this may result in a~loss of~relevant information.

\textbf{Similarity of~tabular datasets.} One way to find structure between such diverse datasets is through the~use of~meta-features~\citep{vanschoren2019meta}. Meta-features account for information about the~tabular datasets' statistical  characteristics. So far, most works have  summarised the~distribution of~each feature in the~dataset. The~basic meta-features are based on statistical definitions (such as kurtosis or skewness)~\citep{Rivolli2018TowardsMeta}, the~performance of~simple machine learning models or features extracted from them~\citep{pfahringer2000meta}.  In addition to the~prior defined meta-features, automatic feature extraction models~\citep{edwards2016towards,jomaa2021dataset2vec} are developed to extract noninterpretable meta-features.  However, all available representations of~tabular datasets focus on reflecting the~structure ignoring the~specificity of~the~domain and the semantic meaning of~the~features.

\textbf{Medical (domain) databases.}
In medicine, specialization and specificity of~datasets are very important and researchers pay much attention to high quality of~data. The~richness of~the~available medical domain datasets is very well demonstrated by the~PhysioNet platform~\citep{PhysioNet}, which was created so that potentially sensitive medical data could be shared responsibly.
This allows large datasets such as MIMIC~\citep{johnson2020mimic}, HiRID~\citep{hyland2020early} or GOSSIS~\citep{raffa2022global} to be shared with the community.
What is more, virtually every country is developing BioBanks, potential sources for the~vast amount of~data used in machine learning. One of~the~most popular in the machine learning community is the~UK BioBank~\citep{sudlow2015uk}. Moreover,  medical specialities also provide similar data, e.g., UNOS in transplantology or SEER~\citep{duggan2016surveillance} in oncology. 
However, standards for integrating data from various experiments leave much to be desired. In the~CTSA's National Center for Data to Health (CD2H)\footnote{\url{https://cd2h.org/}}, guidelines are being developed for the~creation, normalization and sharing of~meta-data to support reusability. Yet there is still little attention paid to giving semantic meaning to individual features and leveraging this knowledge in meta-learning.

\subsection{Ontologies}
\label{sec:rel_onto}

Domain datasets often describe very similar concepts, and to fully comprehend differences we need to understand nuances between them. Often various concepts are described at different levels of~generality. In one dataset we have a~more precise definition of~a~variable while in another we don't have full, detailed information and have to use a~broader term. An example is "blood pressure" and "non-invasive systolic blood pressure". Encoding this hierarchy of~semantic meaning of~individual features requires using an appropriate knowledge base -- an ontology. 

In simple words, an ontology may be understood as a~graph of~terms represented as vertices and edges defining relations between them~\citep{gruber1993translation}.  Usually, ontologies are directed graphs in which a~hierarchy of~concepts can be specified; from more general concepts to specific ones. Thus, ontology enables the~storage of~domain-specific vocabulary and provides means to describe phenomena within a~particular domain in a~format that can be understood and used by computers.

Every domain has its own knowledge bases, with terms and relations reflecting the~specificity of~the~domain. In the~medical domain, there are plenty of~ontologies such as Gene Ontology (GO) \citep{ashburner2000gene, gene2021gene}, Human Phenotype Ontology (HPO) \citep{kohler2021human} or Systematized Nomenclature of~Medicine - Clinical Terms (SNOMED-CT) \citep{wang2002snomed}. Ontologies often encode thousands of~different terms, so the~annotated set of~specific concepts forms a~sparse subset of~all terms. The~semantic similarity between terms is defined to assess the~semantic proximity of~terms considering primarily taxonomic relationships~\citep{harispe2015semantic}.

Semantic similarity allows us to grasp the~proximity between concepts by looking at the~graph structure of~the~ontology and the~information content. Because this may be differently defined, it is challenging to provide an unambiguous definition of~semantic similarity as a~formal measure~\citep{seki2008gene, mcinnes2015evaluating,harispe2014framework, blanchard2008generic}. It~is usually assumed that a~measure is better the~more similar its output would be to the~experts' assessment of~similarity \citep{seki2008gene, harispe2014framework}. \citet{pedersen2007measures} proposed a~benchmark that can be used to evaluate the~correlation between the~similarities returned by measures and those determined by domain experts. The~choice of~semantic similarity depends on the~application and ontology.

Measures of~semantic similarity are successfully used in disciplines such as natural language processing \citep{JIANG2020103581}, geoinformatics \citep{ballatore2013geographic} or even neuroscience \citep{kocon2021mapping}. Among other fields, wide applications have been found for it in the~biomedical domain, where it has been used to compare biological entities by meanings \citep{zhang2016protein, gottlieb2011predict}. In this work, we apply this meaning to structure and represent semantic nets of~features and datasets.

\section{Semantic Feature Net (SeFNet)}

In this section, we present Semantic Feature Net -- an approach introducing semantic  knowledge between tabular datasets. SeFNet serves as a~system that structures a~collection of~features originating from the~considered domain and used in the~machine learning process. This is a universal approach, applicable to many domains. 

To define SeFNet and adapt it to a~specific application, we need to specify three essential components. The~first is \textbf{a set of~tabular datasets} from the selected domain. These datasets serve as the~basis for extracting and structuring the~features. The~second is \textbf{an ontology} that covers the~relevant concepts from the~considered domain and the~datasets. The~choice of~ontology is the~responsibility of~domain experts. The~last one is \textbf{a semantic similarity measure} consistent with the~selected ontology.

After defining the~key components, the~first step of~building SeFNet resource is feature annotation, which produces the~mapping of~features found in datasets to terms in the~selected ontology. This process can be done manually, preferably with the~support of~a~domain expert, but in the future, it~is possible to automate this process. In~the~SeFNet framework, we assume that the~same ontology  is used for annotations of~each dataset. In~this step, the~features gain representation in the~domain knowledge graph.

SeFNet is a~high-level system dependent on the~selections made for datasets and ontologies. We present its components using healthcare datasets as an example. We refer to this repository in the~following sections as SeFNet-Healthcare.

\subsection{Datasets in SeFNet-Healthcare}

{\small
\begin{table}[]
\caption{A summary of~the~annotated datasets and their origins. Each was assigned to one of~two categories. We also provide the~number of~unique variables in each dataset (No.Feat.) and the~number of~annotated features with terms from SNOMED-CT.}
\label{tab:all_df}

\begin{tabular}{llllrrr}
\toprule
 \textbf{ID}      & \textbf{Dataset}               & \textbf{Origin}         & \textbf{Cat.} & \textbf{No.Feat.} & \textbf{No.Ann.}  \\
   \midrule
1  & Cardiovascular Study           & {[}Kaggle{]}            & Survey            & 16                & 15                  \\
2  & Diagnosis of~COVID-19 (Subset) & {[}Kaggle{]}            & EHR               & 19                & 18                  \\
3  & Diabetes Health Indicators     & {[}Kaggle{]}            & Survey            & 22                & 21                  \\
4  & Diabetes 130 US                & {[}UCI,OpenML,Kaggle{]} & EHR               & 49                & 38                  \\
5  & GOSSIS-1-eICU Model Ready      & {[}PhysioNet{]}         & EHR               & 68                & 60                  \\
6  & Stroke Prediction              & {[}Kaggle{]}            & Survey            & 11                & 11                   \\
7  & Heart Disease Indicators       & {[}Kaggle{]}            & Survey            & 22                & 21                  \\
8  & Heart Disease (Comprehensive)  & {[}OpenML{]}            & EHR               & 12                & 11                  \\
9  & HCV data                       & {[}UCI,OpenML,Kaggle{]} & EHR               & 13                & 13                   \\
10 & Hepatitis                      & {[}UCI,Kaggle{]}        & EHR               & 20                & 19                  \\
11 & HiRID Preprocessed             & {[}PhysioNet{]}         & EHR               & 18                & 17                  \\
12 & Pima Indians Diabetes          & {[}OpenML,Kaggle{]}     & EHR               & 9                 & 8                   \\
13 & ILPD                           & {[}UCI,OpenML,Kaggle{]} & EHR               & 11                & 11                   \\
14 & Breast Cancer                  & {[}UCI,OpenML{]}        & EHR               & 10                & 9                   \\
15 & metaMIMIC                      & {[}Paper{]}             & EHR               & 184               & 175                 \\
16 & Thyroid Disease                & {[}UCI,OpenML,Kaggle{]} & EHR               & 30                & 27    \\
\bottomrule
\end{tabular}
\vspace{-5mm}
\end{table}
}

We  initiate the~prototype of~SeFNet for the healthcare domain with the~specification of~datasets. In~Section~\ref{sec:rel_data}, we present a~diversity of~medical datasets, but due to limited resources, we have to limit the~scope of~the~search.  

Our goal is to systematize the~medical datasets  used in machine learning. We target data describing individual cases, enabling the~identification of~risk factors in medical research. For this reason, we focus mainly on two types of~data sources:
(1) Kaggle, OpenML, and UCI repository, where medical datasets from not always well-verified sources are available for immediate download. These reflect the~resource that the~average machine learning researcher works with; (2) PhysioNet platform where high-volume medical datasets are published and available for credentialed usage. These datasets are commonly used in multi-center retrospective studies. For this data, preprocessing is necessary to transform the~data into a~single plain table. We used preprocessing either prepared by database authors or in other research projects.

We have collected $16$ datasets (see Table~\ref{tab:all_df}), which can be divided  into two groups due to the~type of~features included in each dataset: (1) those based on survey data, (2) a~group of~datasets in which information collected with medical devices (EHR) predominates.

The resources collected within SeFNet-Healthcare can be expanded, however, we aimed to include representative examples of~datasets from various sources. The~overview of~the~presented datasets can be easily accessed on the~provided website \url{https://sefnet.mi2.ai/}.  Detailed descriptions of~datasets are provided in Appendix~\ref{app:datasets}.

{\small
\begin{table}[ht]
\centering
\caption{The most common terms in all datasets, along with examples of~how the~variables describing these terms were originally named. }
\label{tab:examples_terms}
\begin{tabular}{p{2.5cm}%
r%
r%
>{\raggedright\arraybackslash}p{6cm}}
  \toprule
 \textbf{SNOMED-CT term ID } & \textbf{No. datasets}& \textbf{No. unique} & \textbf{Names used in datasets} \\ 
  \midrule
 397669002 &  15 &   3 & \textit{Patient age quantile, age, Age} \\ 
 263495000 &  11 &   4 & \textit{Gender, sex, Sex} \\ 
 73211009 &   6 &   5 & \textit{diabetes, diabetes\_diagnosed, Outcome} \\ 
 359986008 &   5 &   8 & \textit{Total\_Bilirubin, bilirubin, BIL} \\ 
 \multirow{2}{*}{38341003} &   \multirow{2}{*}{5} &   \multirow{2}{*}{4} & \textit{HighBP, prevalentHyp, hypertensive\_diagnosed} \\ 
   \bottomrule
\end{tabular}
\vspace{-2mm}
\end{table}
}

\subsection{SNOMED-CT Ontology}

We employ the~SNOMED-CT ontology to describe medical and demographic concepts encoded in variables. SNOMED-CT is not the~only valid choice of~ontology, but it is supported by its universality and the~numerous works that map other vocabularies onto this ontology~\citep{dhombres2016interoperability,thandi2021mapping,kieft2018mapping}. SNOMED-CT ontology is being actively developed and contains more than 350 thousand terms relating to anatomy and demographic data often attached to patient descriptions~\citep{snomed2021summary}. In many countries, the~SNOMED-CT terminology is becoming the~standard.

\textbf{Annotations.} The~annotation process is challenging. Because of~the~brief and often inconsistent feature names in the~considered data, it is challenging to automate that process without additional resources, so it is prepared by manual annotation. Annotations are based on feature names but also on descriptions emerging from dictionaries provided with the~dataset. The latter is especially important for data from PhysioNet since internal system codes are often used to describe the~variables. We establish consistent criteria for annotating ambiguous terms.   The~SeFNet repository is the~first of~its kind to be systematized and made available. The entire resource of annotations is also available on the repository\footnote{\url{https://github.com/MI2DataLab/SeFNet-Healthcare}}  and can be easily explored in SeFNet website.

SNOMED-CT ontology is well suited to the~task of~annotating medical data used in machine learning; 216 different features are included in the~selected datasets, and up to $92\%$ of~them are annotated. Looking at each dataset separately, we also observe a~high percentage of~variable coverage with terms from the~ontology (see Table~\ref{tab:all_df}). The~lowest percentage of~annotated variables has the~dataset \textit{Diabetes 130 US}. However, the~variables that have no equivalent in SNOMED-CT terms were administrative in nature, such as 
\texttt{admission\_type\_id, discharge\_disposition\_id}. 
It is worth noting that for two datasets it was possible to annotate $100\%$ of~the~variables.
On the~one hand, all of~the~data concern medical problems, but on the~other hand, they touch on different specificities of~diseases, e.g.,~diabetes or Covid-19. Nevertheless, a~significant number of~terms occur in more than one dataset. In Table~\ref{tab:examples_terms}, we present the~most frequently recurring terms and examples of~original variable names found in the~data. Patient age and gender are the~most common, but disease information is also prevalent.

\begin{figure}[!h]
    \centering
          \includegraphics[width=0.95\textwidth]{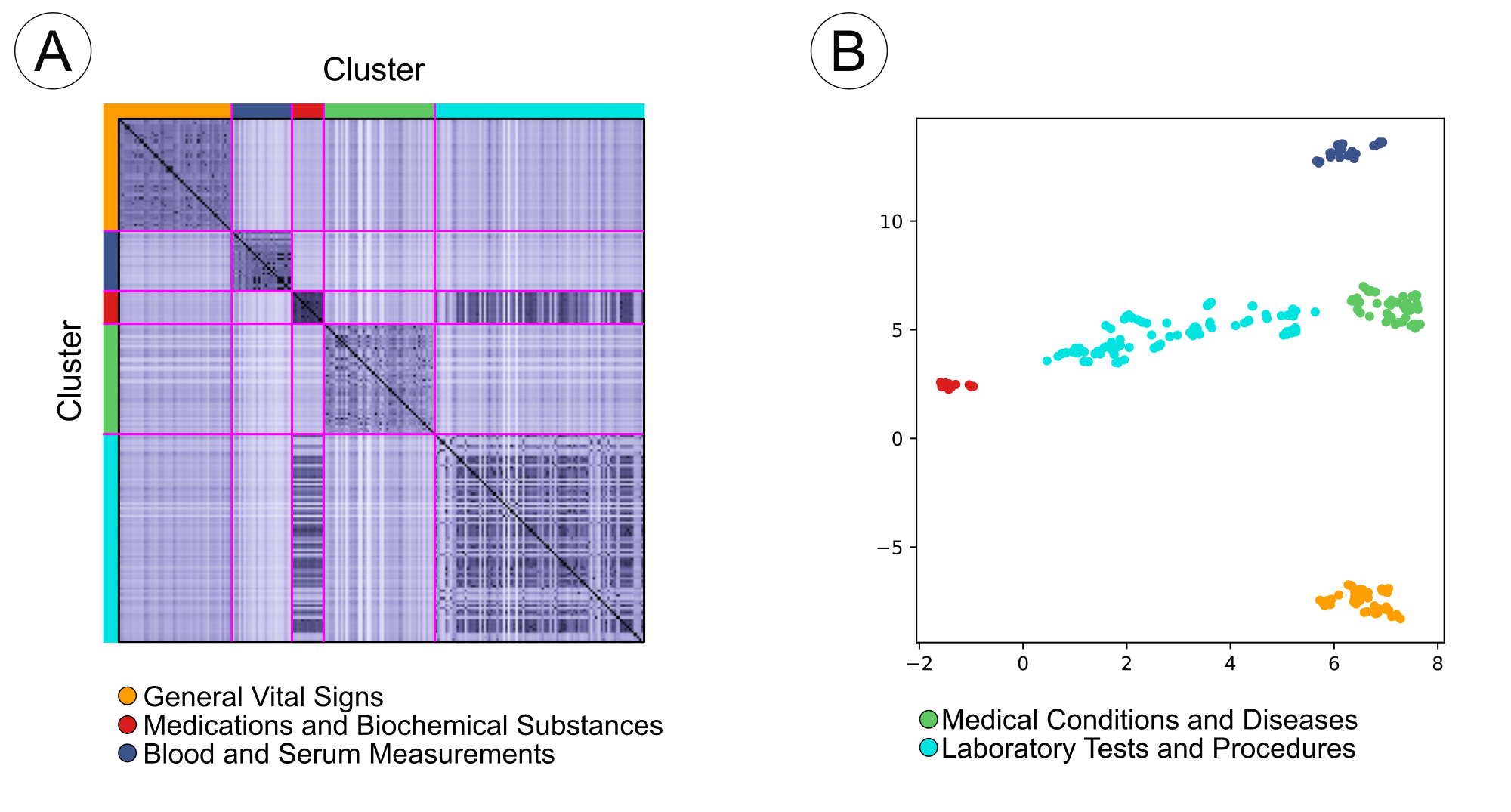}
    \caption{Similarity between the~annotated features. In Panel A, we show the~similarity matrix between each pair of~features. The~features' order corresponds to the~clusters' belonging, illustrated in Panel B. We can distinguish five groups of~features named based on the high-level concepts. 
    }
    \label{fig:sim_terms}
 
\end{figure}

\subsection{Semantic similarity of~terms}

Annotation of~variables allows for the analysis of~datasets regarding jointly occurring variables encoded as terms. However, the~biggest advantage of~using ontologies over dictionaries is using relations between concepts.  As mentioned in Section~\ref{sec:rel_onto}, there is no one universal measure of~similarity, but they are application-dependent. In the~case of~SNOMED-CT,  \citet{harispe2014framework} conducted a~benchmark on 29 terms derived from that ontology and found that the~most concurrent measure with expert intuition is Tversky's abstract Ratio Model measure \citep{tversky1977features} with specific values of~parameters. 

The~measure is formulated in the~sense of~common and distinctive information contained in terms. Let $A(t)$ be a~ set consisting of~a~term $t$ and its ancestors in a~given ontology, and $\Theta(t) = |A(t)|$ be the~cardinality of~this set. Common information between terms $t_1, t_2$ may be encoded as $A(t_1) \cap A(t_2)$, and  we assume that this information is measured as $\Psi(t_1,t_2) = |A(t_1) \cap A(t_2)|$. So we use similarity measure ($SM_{RM}$) between terms $t_1, t_2$ defined as follows:

\begin{equation}
\label{eq:similarity}
    SM_{RM}(t_1,t_2) = \frac{\Theta(t_1)}{ \alpha \cdot \Phi(t_1,t_2) + \beta \cdot \Phi(t_2,t_1) + \Theta(t_1)}, 
\end{equation}
where $\Phi(t_1,t_2) = \Theta(t_1) - \Psi(t_1,t_2)$. Constants $\alpha, \beta$ are determined in \citep{harispe2014framework} so we assume that 
$\alpha=7.9$ and $\beta=3.9$. This measure is symmetric, and values of $(1-SM_{RM})$ can be considered the~distance between terms. Whenever we refer to semantic similarity, we compute it using (\ref{eq:similarity}), as it seems to be fairly effective in determining the~proximity of~terms in the~SNOMED-CT ontology.

In Figure~\ref{fig:sim_terms}A, we show the~values of $SM_{RM}$ for each pair of~terms occurring within SeFNet-Healthcare. Figure~\ref{fig:sim_terms}B visualizes projected vectors of similarity using UMAP~\citep{McInnes2018} technique.  To define the groups of features, we use HDBSCAN~\citep{campello2013density} and then  name them based on high-level concepts in the ontology. The~determined clusters also form a~distinct box structure on the~similarity matrix in panel A.

\section{Dataset Ontology-based Semantic Similarity (DOSS)}

The final contribution of~this paper is  a~method that summarizes semantic similarity between datasets.  In this section, we introduce a~Dataset Ontology-based Semantic Similarity (DOSS) measure that aggregates the~similarity between two sets of~terms in particular between sets of~variables contained in two datasets.

\begin{figure}[!ht]
    \centering
    \includegraphics[scale=0.35]{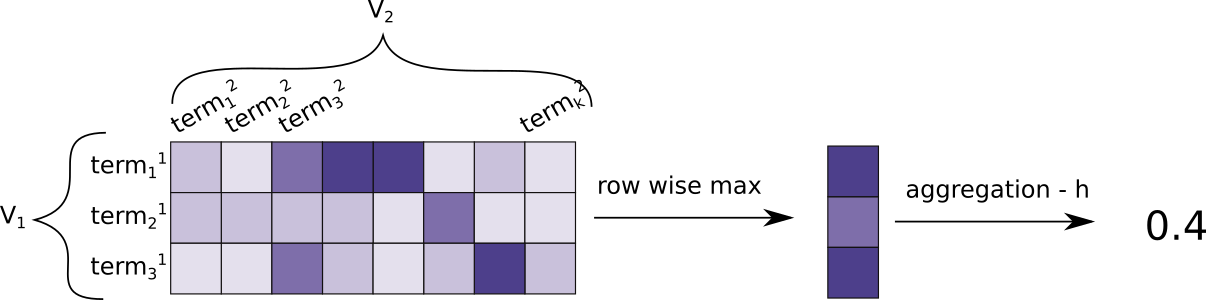}
    \caption{An illustrative scheme for calculating DOSS between $D_1$ and $D_2$. Sets $V_1$ and $V_2$ contains features mapped to $\{term_1^1, \ldots, term^1_m\}$ and $\{term_1^2, \ldots, term^2_k\}$ respectively. Function $h$ is aggregating function.  }
    \label{fig:schema_doss}
\end{figure}

Let $V_1$ and $V_2$ be two sets of~terms included in datasets $D_1, D_2$ respectively.   
 Let each set consist of~a~set of~terms 
 $V_1 = \{ t_1^1, \ldots, t^1_m \} $ , 
 $V_2 = \{ t_1^2, \ldots,  t^2_n \} $. Then we define DOSS as 

\begin{equation}
   DOSS(D_1|D_2) = h(\{ \max_{t_k^2 \in V_2} \;  SM_{RM}(t_i^1, t_k^2) | i \in \{1, \ldots, m \} \}),
\end{equation}

where $h: [0,1]^n \to [0,1]$  is discretionary summarising function. 

\begin{wrapfigure}{r}{0.5\textwidth}
  \begin{center}
   \includegraphics[width=0.45\textwidth]{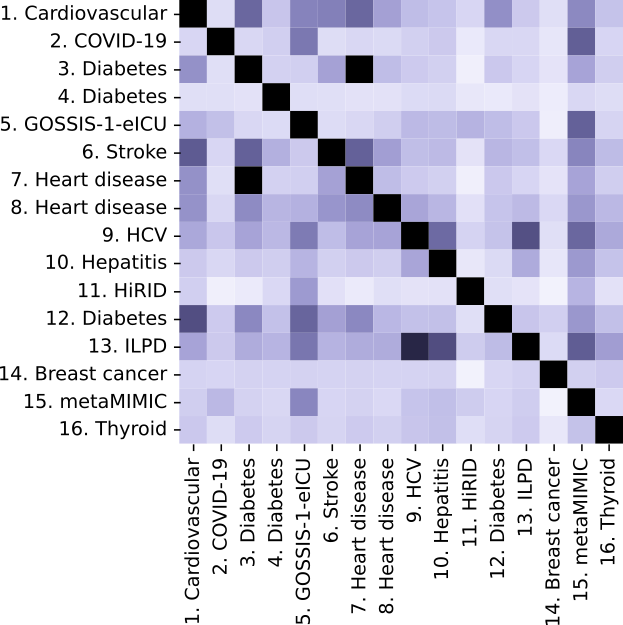}
  \end{center}
  \caption{Matrix of DOSS values between any two annotated datasets. }
  \label{fig:sim_datasets}
\end{wrapfigure}

The mean function is the~most intuitive choice of~$h$ since it determines the~average similarity  of~$V_1$ terms relative to $V_2$. An overview of~the~DOSS calculation scheme is shown in Figure~\ref{fig:schema_doss}.  This measure is not symmetric - we define the~similarity of~one set relative to another.  In case of~similar cardinality of~ $V_1$ and $V_2$  this measure is close to symmetric, but if one set is much bigger than the~second, then the~difference is significant - for instance, $V_2 \subseteq V_1$, the~$DOSS(D_2|D_1) = 1$ but $DOSS(D_1|D_2) < 1$.

We apply the~proposed DOSS measure to the~SeFNet-Healthcare resource. Figure~\ref{fig:sim_datasets} indicates the~values for each pair of~datasets. This matrix is not symmetric,  the~similarity of~smaller datasets (containing small number of~features) to large datasets is higher than in the~opposite case. What is more, the~DOSS similarity  is highly correlated with a number of shared features between two datasets.  However, DOSS is more robust when dealing with terms that are close in meaning but not exactly the same even i.e., if the~intersection of~a~feature set is empty, the~DOSS of~such datasets is greater than zero.

\section{Possible applications}

The introduced SeFNet opens many new opportunities in meta-learning for tabular datasets. In this section, we highlight two specific areas where these opportunities are particularly relevant.

\textbf{Meta-learning.}  
By introducing the structure of features and assessing similarity, SeFNet enables the~exploration of~semantic representations for tabular datasets.  It is important in meta-learning where  we search for descriptive meta-features to assess the potential of transferring information.  The~semantic information embedded in DOSS can enhance the~pool of~existing meta-features (see~Section~\ref{sec:rel_work}) or be used as a~standalone representation. For example, in hyperparameter optimization, DOSS between datasets can serve as weights for tasks from meta-train set to pay more attention to experiments containing features close to these in meta-test tasks. 
Another semantic representation is~to create the~embedding of~datasets. One of~the~most straightforward methods to do that is to generate a~vector representing the~distances between individual features in the~dataset and a~set of~expert terms. These expert terms can be derived from a~meta-test dataset, for example, when not all features in the~target dataset are equally relevant, and we want to give more importance to some of~them. Expert terms can also be specified a~priori in domain expertise.
Dataset representations built on the~basis of~SeFNet designed for a~given class of~domain problems are a~response to previous attempts to add information about feature meaning for a~heterogeneous feature space~\citep{iwata2022sharing}.

\textbf{Incorporating domain knowledge in the~machine learning process.} Integration of~ontologies in SeFNet may be a valuable support to data science specialists, addressing the~challenge of~limited domain knowledge. Although SeFNet does not replace the need to consult experts, it can facilitate communication with them. Designing SeFNet may be beneficial to summarise information about previously conducted experiments, applied methods and their performance. Exploration of~SeFNet structure and~annotated datasets allows us to determine the~importance of~specific variables and~ascertain whether they exist in our dataset. Furthermore, the~ability to identify correlated and~interacting features enables their effective utilization during preprocessing. For example, a~noteworthy application is the~conditional imputation of~missing data, where biological indicators frequently exhibit correlations with each other. By exploiting these features, a~reliable substitute for one indicator can be found in another, acting as a~proxy.

\section{Discussion}
\label{sec:discussion}
\textbf{Maintenance and  updates.} We hope this work is the~beginning of~a~useful growing project with community feedback and support. New annotated datasets can be uploaded through the~form on the~website. We will review them and if suitable they will expand the collection.

\textbf{Limitations. } In order to ensure the~proper use of~the~individual datasets, we strongly advise researchers to refer to the~official documentation and resources. We must emphasize that we are not the~creators of~the~datasets and, therefore cannot offer any endorsement or guarantees regarding their accuracy.   Due to SNOMED-CT policy, we cannot share the entire structure of this ontology, but we can share information about the terms used.  To gain insight into the graph structure, it is necessary to acquire access to SNOMED-CT.

\textbf{Negative societal impacts.} Our proposed SeFNet resource designed for healthcare uses the~SNOMED-CT ontology, which is not available in all countries.  A~similar problem may appear when using commercial knowledge bases and ontologies.

\section{Conclusions}

In this paper, we introduced an approach SeFNet reflecting  the~semantics of~features found in tabular datasets. To our knowledge, this is the~first work incorporating semantic feature information in the~representation of~tasks based on tabular data.  This additional layer of~information allows us to gain insights into the~semantic structure and may lead to more informed decision-making in designing machine learning solutions for specific tasks. 

A very important aspect of~future work is the~integration of~SeFNet into end-to-end machine learning pipelines. Investigating how SeFNet can be seamlessly incorporated into various stages of~the~machine learning processes, such as feature selection, data imputation, or model optimization, would enhance its practical utility and adoption in real-world applications. 
The next point that can be developed is refining the~methods for quantifying the~semantic similarity between datasets using SeFNet. The~current approach utilizes the~Dataset Ontology-based Semantic Similarity (DOSS) aggregation, but further research can explore alternative techniques.

This project lays the~foundation for feature-centric approaches in machine learning. In the~near future, we plan to further develop the~created repository of~datasets but also to explore the~potential of~automating the~addition of~variable structure information to the~machine-learning process.

\section{Acknowledgments}

The work on this paper is financially supported by the~NCN Sonata Bis-9 grant 2019/34/E/ST6/00052 and  NCBiR grant INFOSTRATEG-I/0022/2021-00. The research were carried out on devices cofunded by the Warsaw University of Technology within the Excellence Initiative: Research University (IDUB) programme.

\bibliographystyle{abbrvnat} 
\bibliography{arxiv/references_arxiv}

\begin{thebibliography}{54}
\providecommand{\natexlab}[1]{#1}
\providecommand{\url}[1]{\texttt{#1}}
\expandafter\ifx\csname urlstyle\endcsname\relax
  \providecommand{\doi}[1]{doi: #1}\else
  \providecommand{\doi}{doi: \begingroup \urlstyle{rm}\Url}\fi

\bibitem[Alaa and Schaar(2018)]{alaa2018autoprognosis}
A.~Alaa and M.~Schaar.
\newblock {AutoPrognosis: Automated Clinical Prognostic Modeling via Bayesian
  Optimization with Structured Kernel Learning}.
\newblock In \emph{Proceedings of the 35th International Conference on Machine
  Learning (ICML)}, pages 139--148, 2018.

\bibitem[Alizadehsani et~al.(2019)Alizadehsani, Roshanzamir, Abdar,
  Beykikhoshk, Khosravi, Panahiazar, Koohestani, Khozeimeh, Nahavandi, and
  Sarrafzadegan]{alizadehsani2019database}
R.~Alizadehsani, M.~Roshanzamir, M.~Abdar, A.~Beykikhoshk, A.~Khosravi,
  M.~Panahiazar, A.~Koohestani, F.~Khozeimeh, S.~Nahavandi, and
  N.~Sarrafzadegan.
\newblock A database for using machine learning and data mining techniques for
  coronary artery disease diagnosis.
\newblock \emph{Scientific data}, 6\penalty0 (1):\penalty0 227, 2019.
\newblock \doi{https://doi.org/10.1038/s41597-019-0206-3}.

\bibitem[Ashburner et~al.(2000)Ashburner, Ball, Blake, Botstein, Butler,
  Cherry, Davis, Dolinski, Dwight, Eppig, et~al.]{ashburner2000gene}
M.~Ashburner, C.~A. Ball, J.~A. Blake, D.~Botstein, H.~Butler, J.~M. Cherry,
  A.~P. Davis, K.~Dolinski, S.~S. Dwight, J.~T. Eppig, et~al.
\newblock Gene ontology: tool for the unification of biology.
\newblock \emph{Nature Genetics}, 25\penalty0 (1):\penalty0 25--29, 2000.
\newblock \doi{https://doi.org/10.1038/s41597-019-0206-3}.

\bibitem[Ballatore et~al.(2013)Ballatore, Bertolotto, and
  Wilson]{ballatore2013geographic}
A.~Ballatore, M.~Bertolotto, and D.~C. Wilson.
\newblock {Geographic knowledge extraction and semantic similarity in
  OpenStreetMap}.
\newblock \emph{Knowledge and Information Systems}, 37\penalty0 (1):\penalty0
  61--81, 2013.
\newblock \doi{https://doi.org/10.1007/s10115-012-0571-0}.

\bibitem[Bischl et~al.(2021)Bischl, Casalicchio, Feurer, Gijsbers, Hutter,
  Lang, Mantovani, van Rijn, and Vanschoren]{bischl2017openml100}
B.~Bischl, G.~Casalicchio, M.~Feurer, P.~Gijsbers, F.~Hutter, M.~Lang, R.~G.
  Mantovani, J.~N. van Rijn, and J.~Vanschoren.
\newblock Open{ML} benchmarking suites.
\newblock In \emph{Proceedings of the 35th Conference on Neural Information
  Processing Systems Datasets and Benchmarks Track (Round 2)}, 2021.
\newblock URL \url{https://openreview.net/forum?id=OCrD8ycKjG}.

\bibitem[Blanchard et~al.(2008)Blanchard, Harzallah, and
  Kuntz]{blanchard2008generic}
E.~Blanchard, M.~Harzallah, and P.~Kuntz.
\newblock A generic framework for comparing semantic similarities on a
  subsumption hierarchy.
\newblock In \emph{Proceedings of the 18th European Conference on Artificial
  Intelligence (ECAI)}, pages 20--24, 2008.

\bibitem[Campello et~al.(2013)Campello, Moulavi, and
  Sander]{campello2013density}
R.~J. Campello, D.~Moulavi, and J.~Sander.
\newblock Density-based clustering based on hierarchical density estimates.
\newblock In \emph{Proceedings of the 17th Pacific-Asia Conference in Knowledge
  Discovery and Data Mining (PAKDD)}, pages 160--172. Springer, 2013.

\bibitem[Chui et~al.(2018)Chui, Manyika, Miremadi, Henke, Chung, Nel, and
  Malhotra]{mckinsey}
M.~Chui, J.~Manyika, M.~Miremadi, N.~Henke, R.~Chung, P.~Nel, and S.~Malhotra.
\newblock {Notes from the AI frontier: insights from hundreds of use cases},
  2018.

\bibitem[Dhombres and Bodenreider(2016)]{dhombres2016interoperability}
F.~Dhombres and O.~Bodenreider.
\newblock {Interoperability between phenotypes in research and healthcare
  terminologies—Investigating partial mappings between HPO and SNOMED CT}.
\newblock \emph{Journal of Biomedical Semantics}, 7:\penalty0 1--13, 2016.
\newblock \doi{10.1186/s13326-016-0047-3}.

\bibitem[Dua and Graff(2017)]{Dua:2019}
D.~Dua and C.~Graff.
\newblock {UCI Machine Learning Repository}, 2017.
\newblock URL \url{http://archive.ics.uci.edu/ml}.

\bibitem[Duggan et~al.(2016)Duggan, Anderson, Altekruse, Penberthy, and
  Sherman]{duggan2016surveillance}
M.~A. Duggan, W.~F. Anderson, S.~Altekruse, L.~Penberthy, and M.~E. Sherman.
\newblock The surveillance, epidemiology and end results (seer) program and
  pathology: towards strengthening the critical relationship.
\newblock \emph{The American Journal of Surgical Pathology}, 40\penalty0
  (12):\penalty0 e94--e102, 2016.
\newblock \doi{10.1097/PAS.0000000000000749}.

\bibitem[Edwards and Storkey(2017)]{edwards2016towards}
H.~Edwards and A.~Storkey.
\newblock {Towards a Neural Statistician}.
\newblock In \emph{Proceedings of the 5th International Conference on Learning
  Representations (ICLR)}, pages 1--13, 2017.

\bibitem[Faltys et~al.(2021)Faltys, Zimmermann, Lyu, Hüser, Hyland, Rätsch,
  and Merz]{faltys2022hirid}
M.~Faltys, M.~Zimmermann, X.~Lyu, M.~Hüser, S.~Hyland, G.~Rätsch, and
  T.~Merz.
\newblock {HiRID, a high time-resolution ICU dataset}.
\newblock PhysioNet, 2021.

\bibitem[Feurer et~al.(2015)Feurer, Springenberg, and
  Hutter]{feurer2015initializing}
M.~Feurer, J.~Springenberg, and F.~Hutter.
\newblock Initializing bayesian hyperparameter optimization via meta-learning.
\newblock In \emph{Proceedings of the 29th AAAI Conference on Artificial
  Intelligence}, volume~29, 2015.
\newblock \doi{https://doi.org/10.1609/aaai.v29i1.9354}.

\bibitem[{Gene Ontology Consortium}(2021)]{gene2021gene}
{Gene Ontology Consortium}.
\newblock {The Gene Ontology resource: enriching a GOld mine}.
\newblock \emph{Nucleic Acids Research}, 49\penalty0 (D1):\penalty0 D325--D334,
  2021.
\newblock \doi{10.1093/nar/gkaa1113}.

\bibitem[Gijsbers et~al.(2022)Gijsbers, Bueno, Coors, LeDell, Poirier, Thomas,
  Bischl, and Vanschoren]{ambl}
P.~Gijsbers, M.~L. Bueno, S.~Coors, E.~LeDell, S.~Poirier, J.~Thomas,
  B.~Bischl, and J.~Vanschoren.
\newblock Amlb: an automl benchmark.
\newblock \emph{arXiv preprint arXiv:2207.12560}, 2022.

\bibitem[Goldberger et~al.(2000)Goldberger, Amaral, Glass, Hausdorff, Ivanov,
  Mark, Mietus, Moody, Peng, and Stanley]{PhysioNet}
A.~L. Goldberger, L.~A.~N. Amaral, L.~Glass, J.~M. Hausdorff, P.~C. Ivanov,
  R.~G. Mark, J.~E. Mietus, G.~B. Moody, C.-K. Peng, and H.~E. Stanley.
\newblock {{PhysioBank, PhysioToolkit, and PhysioNet}: Components of a New
  Research Resource for Complex Physiologic Signals}.
\newblock \emph{Circulation}, 101\penalty0 (23):\penalty0 215--220, 2000.

\bibitem[Gottlieb et~al.(2011)Gottlieb, Stein, Ruppin, and
  Sharan]{gottlieb2011predict}
A.~Gottlieb, G.~Y. Stein, E.~Ruppin, and R.~Sharan.
\newblock {PREDICT: a method for inferring novel drug indications with
  application to personalized medicine}.
\newblock \emph{Molecular Systems Biology}, 7\penalty0 (1):\penalty0 496, 2011.
\newblock \doi{10.1038/msb.2011.26}.

\bibitem[Gruber(1993)]{gruber1993translation}
T.~R. Gruber.
\newblock A translation approach to portable ontology specifications.
\newblock \emph{Knowledge Acquisition}, 5\penalty0 (2):\penalty0 199--220,
  1993.
\newblock \doi{10.1006/KNAC.1993.1008}.

\bibitem[Harispe et~al.(2014)Harispe, S{\'a}nchez, Ranwez, Janaqi, and
  Montmain]{harispe2014framework}
S.~Harispe, D.~S{\'a}nchez, S.~Ranwez, S.~Janaqi, and J.~Montmain.
\newblock A framework for unifying ontology-based semantic similarity measures:
  A study in the biomedical domain.
\newblock \emph{Journal of Biomedical Informatics}, 48:\penalty0 38--53, 2014.
\newblock \doi{10.1016/j.jbi.2013.11.006}.

\bibitem[Harispe et~al.(2015)Harispe, Ranwez, Janaqi, and
  Montmain]{harispe2015semantic}
S.~Harispe, S.~Ranwez, S.~Janaqi, and J.~Montmain.
\newblock Semantic similarity from natural language and ontology analysis.
\newblock \emph{Synthesis Lectures on Human Language Technologies}, 8\penalty0
  (1):\penalty0 1--254, 2015.
\newblock \doi{10.2200/S00639ED1V01Y201504HLT027}.

\bibitem[Hutter et~al.(2019)Hutter, Kotthoff, and
  Vanschoren]{hutter2019automated}
F.~Hutter, L.~Kotthoff, and J.~Vanschoren.
\newblock \emph{Automated Machine Learning: Methods, Systems, Challenges}.
\newblock Springer Nature, 2019.
\newblock ISBN 978-3-030-05317-8.

\bibitem[Hyland et~al.(2020)Hyland, Faltys, H{\"u}ser, Lyu, Gumbsch, Esteban,
  Bock, Horn, Moor, Rieck, et~al.]{hyland2020early}
S.~L. Hyland, M.~Faltys, M.~H{\"u}ser, X.~Lyu, T.~Gumbsch, C.~Esteban, C.~Bock,
  M.~Horn, M.~Moor, B.~Rieck, et~al.
\newblock Early prediction of circulatory failure in the intensive care unit
  using machine learning.
\newblock \emph{Nature Medicine}, 26\penalty0 (3):\penalty0 364--373, 2020.
\newblock \doi{https://doi.org/10.1038/s41591-020-0789-4}.

\bibitem[Iwata and Kumagai(2020)]{iwata2020meta}
T.~Iwata and A.~Kumagai.
\newblock Meta-learning from tasks with heterogeneous attribute spaces.
\newblock \emph{Advances in Neural Information Processing Systems},
  33:\penalty0 6053--6063, 2020.

\bibitem[Iwata and Kumagai(2022)]{iwata2022sharing}
T.~Iwata and A.~Kumagai.
\newblock Sharing knowledge for meta-learning with feature descriptions.
\newblock \emph{Advances in Neural Information Processing Systems},
  35:\penalty0 16637--16649, 2022.

\bibitem[Jiang et~al.(2020)Jiang, Wu, Tomita, Ganoe, and
  Hassanpour]{JIANG2020103581}
S.~Jiang, W.~Wu, N.~Tomita, C.~Ganoe, and S.~Hassanpour.
\newblock {Multi-Ontology Refined Embeddings (MORE): A hybrid multi-ontology
  and corpus-based semantic representation model for biomedical concepts}.
\newblock \emph{Journal of Biomedical Informatics}, 111:\penalty0 103581, 2020.
\newblock \doi{https://doi.org/10.1016/j.jbi.2020.103581}.

\bibitem[Johnson et~al.(2020)Johnson, Bulgarelli, Pollard, Horng, Celi, and
  Mark]{johnson2020mimic}
A.~Johnson, L.~Bulgarelli, T.~Pollard, S.~Horng, L.~Celi, and R.~Mark.
\newblock {MIMIC-IV (version 1.0)}, 2020.

\bibitem[Johnson et~al.(2023)Johnson, Bulgarelli, Shen, Gayles, Shammout,
  Horng, Pollard, Moody, Gow, Lehman, et~al.]{johnson2023mimic}
A.~E. Johnson, L.~Bulgarelli, L.~Shen, A.~Gayles, A.~Shammout, S.~Horng, T.~J.
  Pollard, B.~Moody, B.~Gow, L.-w.~H. Lehman, et~al.
\newblock Mimic-iv, a freely accessible electronic health record dataset.
\newblock \emph{Scientific data}, 10\penalty0 (1):\penalty0 1, 2023.

\bibitem[Jomaa et~al.(2021)Jomaa, Schmidt-Thieme, and
  Grabocka]{jomaa2021dataset2vec}
H.~S. Jomaa, L.~Schmidt-Thieme, and J.~Grabocka.
\newblock {Dataset2Vec: learning dataset meta-features}.
\newblock \emph{Data Mining and Knowledge Discovery}, 35\penalty0 (3):\penalty0
  964--985, feb 2021.
\newblock \doi{10.1007/s10618-021-00737-9}.

\bibitem[Kieft et~al.(2018)Kieft, Vreeke, De~Groot, de~Graaf-Waar, van Gool,
  Koster, ten Napel, Francke, and Delnoij]{kieft2018mapping}
R.~Kieft, E.~M. Vreeke, E.~De~Groot, H.~de~Graaf-Waar, C.~H. van Gool,
  N.~Koster, H.~ten Napel, A.~L. Francke, and D.~M. Delnoij.
\newblock {Mapping the Dutch SNOMED CT subset to Omaha system, NANDA
  international and international classification of functioning, disability and
  health}.
\newblock \emph{International Journal of Medical Informatics}, 111:\penalty0
  77--82, 2018.
\newblock \doi{10.1016/j.ijmedinf.2017.12.025}.

\bibitem[Koco{\'n} and Maziarz(2021)]{kocon2021mapping}
J.~Koco{\'n} and M.~Maziarz.
\newblock {Mapping WordNet onto human brain connectome in emotion processing
  and semantic similarity recognition}.
\newblock \emph{Information Processing \& Management}, 58\penalty0
  (3):\penalty0 102530, 2021.
\newblock \doi{https://doi.org/10.1016/j.ipm.2021.102530}.

\bibitem[K{\"o}hler et~al.(2021)K{\"o}hler, Gargano, Matentzoglu, Carmody,
  Lewis-Smith, Vasilevsky, Danis, Balagura, Baynam, Brower,
  et~al.]{kohler2021human}
S.~K{\"o}hler, M.~Gargano, N.~Matentzoglu, L.~C. Carmody, D.~Lewis-Smith, N.~A.
  Vasilevsky, D.~Danis, G.~Balagura, G.~Baynam, A.~M. Brower, et~al.
\newblock {The Human Phenotype Ontology in 2021}.
\newblock \emph{Nucleic Acids Research}, 49\penalty0 (D1):\penalty0
  D1207--D1217, 2021.
\newblock \doi{https://doi.org/10.1093/nar/gkaa1043}.

\bibitem[Lichtinghagen et~al.(2013)Lichtinghagen, Pietsch, Bantel, Manns,
  Brand, and Bahr]{lichtinghagen2013enhanced}
R.~Lichtinghagen, D.~Pietsch, H.~Bantel, M.~P. Manns, K.~Brand, and M.~J. Bahr.
\newblock The enhanced liver fibrosis (elf) score: normal values, influence
  factors and proposed cut-off values.
\newblock \emph{Journal of hepatology}, 59\penalty0 (2):\penalty0 236--242,
  2013.

\bibitem[McInnes and Pedersen(2015)]{mcinnes2015evaluating}
B.~T. McInnes and T.~Pedersen.
\newblock Evaluating semantic similarity and relatedness over the semantic
  grouping of clinical term pairs.
\newblock \emph{Journal of Biomedical Informatics}, 54:\penalty0 329--336,
  2015.
\newblock \doi{https://doi.org/10.1016/j.jbi.2014.11.014}.

\bibitem[McInnes et~al.(2018)McInnes, Healy, Saul, and
  Großberger]{McInnes2018}
L.~McInnes, J.~Healy, N.~Saul, and L.~Großberger.
\newblock Umap: Uniform manifold approximation and projection.
\newblock \emph{Journal of Open Source Software}, 3\penalty0 (29):\penalty0
  861, 2018.
\newblock \doi{10.21105/joss.00861}.

\bibitem[Pedersen et~al.(2007)Pedersen, Pakhomov, Patwardhan, and
  Chute]{pedersen2007measures}
T.~Pedersen, S.~V. Pakhomov, S.~Patwardhan, and C.~G. Chute.
\newblock Measures of semantic similarity and relatedness in the biomedical
  domain.
\newblock \emph{Journal of Biomedical Informatics}, 40\penalty0 (3):\penalty0
  288--299, 2007.
\newblock \doi{https://doi.org/10.1016/j.jbi.2006.06.004}.

\bibitem[Pfahringer et~al.(2000)Pfahringer, Bensusan, and
  Giraud-Carrier]{pfahringer2000meta}
B.~Pfahringer, H.~Bensusan, and C.~G. Giraud-Carrier.
\newblock {Meta-Learning by Landmarking Various Learning Algorithms}.
\newblock In \emph{Proceedings of the 17th International Conference on Machine
  Learning (ICML)}, pages 743--750, 2000.

\bibitem[Raffa et~al.(2022{\natexlab{a}})Raffa, Johnson, Pollard, and
  Omar]{raffa2022gossis}
J.~Raffa, A.~Johnson, T.~Pollard, and B.~Omar.
\newblock {GOSSIS-1-eICU, the eICU-CRD subset of the Global Open Source
  Severity of Illness Score (GOSSIS-1) dataset (version 1.0.0)}.
\newblock PhysioNet, 2022{\natexlab{a}}.

\bibitem[Raffa et~al.(2022{\natexlab{b}})Raffa, Johnson, O’Brien, Pollard,
  Mark, Celi, Pilcher, and Badawi]{raffa2022global}
J.~D. Raffa, A.~E. Johnson, Z.~O’Brien, T.~J. Pollard, R.~G. Mark, L.~A.
  Celi, D.~Pilcher, and O.~Badawi.
\newblock {The global open source severity of illness score (GOSSIS)}.
\newblock \emph{Critical Care Medicine}, 50\penalty0 (7):\penalty0 1040--1050,
  2022{\natexlab{b}}.
\newblock \doi{10.1097/CCM.0000000000005518}.

\bibitem[Ramana et~al.(2011)Ramana, Babu, Venkateswarlu,
  et~al.]{ramana2011critical}
B.~V. Ramana, M.~S.~P. Babu, N.~Venkateswarlu, et~al.
\newblock A critical study of selected classification algorithms for liver
  disease diagnosis.
\newblock \emph{International Journal of Database Management Systems},
  3\penalty0 (2):\penalty0 101--114, 2011.
\newblock \doi{10.5121/IJDMS.2011.3207}.

\bibitem[Rivolli et~al.(2022)Rivolli, Garcia, Soares, Vanschoren, and
  de~Carvalho]{Rivolli2018TowardsMeta}
A.~Rivolli, L.~P. Garcia, C.~Soares, J.~Vanschoren, and A.~C. de~Carvalho.
\newblock Meta-features for meta-learning.
\newblock \emph{Knowledge-Based Systems}, 240:\penalty0 108101, 2022.
\newblock \doi{https://doi.org/10.1016/j.knosys.2021.108101}.

\bibitem[Seki and Mostafa(2008)]{seki2008gene}
K.~Seki and J.~Mostafa.
\newblock Gene ontology annotation as text categorization: An empirical study.
\newblock \emph{Information Processing \& Management}, 44\penalty0
  (5):\penalty0 1754--1770, 2008.
\newblock \doi{https://doi.org/10.1016/j.ipm.2008.05.003}.

\bibitem[{SNOMED~International}(2021)]{snomed2021summary}
{SNOMED~International}.
\newblock Snomed: Executive summary, 2021.
\newblock
  \url{https://www.snomed.org/_files/ugd/900274_8a849a3565054d14a4c94cf1062331a3.pdf}.

\bibitem[Strack et~al.(2014)Strack, DeShazo, Gennings, Olmo, Ventura, Cios, and
  Clore]{strack2014impact}
B.~Strack, J.~P. DeShazo, C.~Gennings, J.~L. Olmo, S.~Ventura, K.~J. Cios, and
  J.~N. Clore.
\newblock {Impact of HbA1c measurement on hospital readmission rates: analysis
  of 70,000 clinical database patient records}.
\newblock \emph{BioMed research international}, 2014.
\newblock \doi{10.1155/2014/781670}.

\bibitem[Sudlow et~al.(2015)Sudlow, Gallacher, Allen, Beral, Burton, Danesh,
  Downey, Elliott, Green, Landray, et~al.]{sudlow2015uk}
C.~Sudlow, J.~Gallacher, N.~Allen, V.~Beral, P.~Burton, J.~Danesh, P.~Downey,
  P.~Elliott, J.~Green, M.~Landray, et~al.
\newblock {UK biobank: an open access resource for identifying the causes of a
  wide range of complex diseases of middle and old age}.
\newblock \emph{PLoS medicine}, 12\penalty0 (3):\penalty0 e1001779, 2015.
\newblock \doi{10.1371/journal.pmed.1001779}.

\bibitem[Thandi et~al.(2021)Thandi, Brown, and Wong]{thandi2021mapping}
M.~Thandi, S.~Brown, and S.~T. Wong.
\newblock Mapping frailty concepts to snomed ct.
\newblock \emph{International Journal of Medical Informatics}, 149:\penalty0
  104409, 2021.
\newblock \doi{https://doi.org/10.1016/j.ijmedinf.2021.104409}.

\bibitem[Tversky(1977)]{tversky1977features}
A.~Tversky.
\newblock Features of similarity.
\newblock \emph{Psychological Review}, 84\penalty0 (4):\penalty0 327–352,
  1977.
\newblock \doi{https://doi.org/10.1037/0033-295X.84.4.327}.

\bibitem[Vanschoren(2019)]{vanschoren2019meta}
J.~Vanschoren.
\newblock \emph{Meta-Learning}, pages 35--61.
\newblock Springer International Publishing, Cham, 2019.
\newblock \doi{10.1007/978-3-030-05318-5_2}.

\bibitem[Vanschoren et~al.(2013)Vanschoren, van Rijn, Bischl, and
  Torgo]{OpenML2013}
J.~Vanschoren, J.~N. van Rijn, B.~Bischl, and L.~Torgo.
\newblock Openml: networked science in machine learning.
\newblock \emph{SIGKDD Explorations}, 15\penalty0 (2):\penalty0 49--60, 2013.
\newblock \doi{10.1145/2641190.2641198}.

\bibitem[Wang et~al.(2002)Wang, Sable, and Spackman]{wang2002snomed}
A.~Y. Wang, J.~H. Sable, and K.~A. Spackman.
\newblock {The SNOMED clinical terms development process: refinement and
  analysis of content.}
\newblock In \emph{Proceedings of the American Medical Informatics Association
  Symposium (AMIA)}, page 845. American Medical Informatics Association, 2002.

\bibitem[Wistuba et~al.(2016)Wistuba, Schilling, and
  Schmidt-Thieme]{Wistuba2016SequentialTuning}
M.~Wistuba, N.~Schilling, and L.~Schmidt-Thieme.
\newblock {Sequential model-free hyperparameter tuning}.
\newblock In \emph{Proceedings in the IEEE International Conference on Data
  Mining (ICDM)}, pages 1033--1038, 2016.
\newblock \doi{10.1109/ICDM.2015.20}.

\bibitem[Wo{\'z}nica et~al.(2022)Wo{\'z}nica, Grzyb, Trafas, and
  Biecek]{woznica2022consolidated}
K.~Wo{\'z}nica, M.~Grzyb, Z.~Trafas, and P.~Biecek.
\newblock Consolidated learning--a domain-specific model-free optimization
  strategy with examples for xgboost and mimic-iv.
\newblock \emph{arXiv preprint arXiv:2201.11815}, 2022.

\bibitem[Xie et~al.(2019)Xie, Nikolayeva, Luo, and Li]{xie2019peer}
Z.~Xie, O.~Nikolayeva, J.~Luo, and D.~Li.
\newblock Building risk prediction models for type 2 diabetes using machine
  learning techniques.
\newblock \emph{Preventing Chronic Disease}, 16, 2019.
\newblock \doi{doi: 10.5888/pcd16.190109}.

\bibitem[Zhang and Tang(2016)]{zhang2016protein}
S.-B. Zhang and Q.-R. Tang.
\newblock Protein--protein interaction inference based on semantic similarity
  of gene ontology terms.
\newblock \emph{Journal of Theoretical Biology}, 401:\penalty0 30--37, 2016.

\end{thebibliography}

\appendix

\section{Datasets}
\label{app:datasets}
\subsection{Survey Study}
\begin{itemize}[leftmargin=0.2in]
  \setlength\itemsep{0.1em}
\item \textit{Cardiovascular Study.} Open source data from a~cardiovascular study on residents of~Framingham, Massachusetts. The~classification goal is to assess the~10-year patient's risk of~future coronary heart disease (CHD). The~dataset contains 4 238 instances and 16 variables, including demographic data, survey information, and a~few EHR-based fields. 

\item \textit{BRFSS Health Indicators.} The~dataset includes 253 680 survey responses from the~Behavioral Risk Factor Surveillance System\footnote{\url{https://www.cdc.gov/brfss/index.html}} (BRFSS) from 2015. This is an example of~an annually collected health-related telephone survey published since 1984. The~original data describes over 300 variables, but the~cleaned version contains 22 features. BRFSS data are utilized in many research~\citep{xie2019peer}. In that experiment, the~objective was to predict the~occurrence of~2-type diabetes (\textit{BRFSS Diabetes Health Indicators}) or the~occurrence of~heart diseases (\textit{BRFSS Heart Disease Indicators}).

\item \textit{Stroke Prediction.} The~dataset describes 5 110 instances with 12 variables used to predict whether a~patient is likely to get a~stroke. This data has no credentialed resources and was made available for educational purposes. Variables include gender, age, information about comorbidities, and smoking status. 

\end{itemize}
\subsection{EHR open data}

First, we discuss datasets made available under unrestricted access. These datasets are very often used for educational purposes and are available within OpenML or the~Kaggle platform.
ok
\begin{itemize}[leftmargin=0.2in]
  \setlength\itemsep{0.1em}
    \item \textit{Diabetes 130 US.} The~dataset represents EHR results saved for ten years (1999-2008) in clinical care units at 130 US hospitals and integrated delivery networks. Originally data comes from~\citep{strack2014impact}. Data includes 101 766 observations.  Variables contain a~description of~the~patient's condition at the~time of~admission, information about the~diagnosis, and the~number of~tests performed.

\item \textit{Diagnosis of~COVID-19.} The~dataset contains anonymized information about patients admitted at the~Hospital Israelita Albert Einstein in São Paulo, Brazil. The~goal of~admission was to perform the~SARS-CoV-2 RT-PCR. Next to that also, additional laboratory tests were performed during a~visit to the~hospital. The~dataset was published in 2020.

\item \textit{Heart Disease (Comprehensive).}	This dataset is curated by combining five datasets over 11 standard features, making it the~largest heart disease dataset available for research. Despite sharing this data on OpenML, it comes from separate research studies and is merged as a~result of~the~meta-analysis ~\citep{alizadehsani2019database}.

\item \textit{HCV data.}	The~dataset contains results for 615 patients collected by~\citet{lichtinghagen2013enhanced}.	Observed patients are blood donors and Hepatitis C patients. Demographic features like age are reported next to laboratory results.

\item \textit{Hepatitis.}	Data for mortality prediction among patients with hepatitis symptoms, including fatigue, anorexia, or big liver. As EHR results, we consider information about albumin and bilirubin level. This dataset is available mostly for educational purposes and has been employed in machine learning research since the~2000s.

\item \textit{Pima Indians Diabetes.} Originally, the~dataset came from the~National Institute of~Diabetes and Digestive and Kidney Diseases, but data was restricted because of~ethical guidelines. The~objective of~the~experiment is to predict whether a~patient has diabetes based on certain diagnostic measurements. This dataset is one of~the~most popular data used to introduce machine learning methods.

\item \textit{Breast Cancer.}	This dataset includes 286 instances described by nine attributes, including categorical features. This is an example of imbalanced data. The~goal of corresponding predictive task is to predict the~occurrence of~breast cancer.

\item \textit{ILPD.} The~dataset was~collected  to detect patients with liver disease~\citep{ramana2011critical}. Data comes from Andhra Pradesh in India. This~dataset contains information about 583 patients and 11 variables.

\item \textit{Thyroid Disease.}		This dataset was created by combining 6 different sources. All of them were collected in Australia.  The dataset is used to identify prognostic factors in thyroid disease  among 30 different features. These include information from blood tests but also from the patient's interview.

\end{itemize}

\subsection{EHR credentialed data}

The medical data used in more advanced studies are not so easily accessible due to the~possibility of~de-anonymizing the~data. We collected information on three datasets from the~PhysioNet platform:

\begin{itemize}[leftmargin=0.2in]
  \setlength\itemsep{0.1em}

\item  \textit{GOSSIS-1-eICU. } Data are collected in the~project including the~subset of~patients in the~USA derived from the~eICU Collaborative Research Database (eICU-CRD) as Global Open Source Severity of~Illness Score~\citep{raffa2022global, raffa2022gossis}. The~dataset consists of~the~information reported within the~first 24 hours after admission for 131 thousand unique patients from 204 hospitals from ICU admissions discharged in 2014-15.  

\item \textit{HiRID.}	High time-resolution ICU dataset is a~freely accessible critical care dataset containing data from~almost 34 thousand patients admitted to the~Department of~Intensive Care Medicine (ICU) of~the~Bern University Hospital in  Switzerland ~\citep{hyland2020early, faltys2022hirid}. HiRID has a~high time resolution of~registered data, most importantly for bedside monitoring, with most parameters recorded every 2 minutes. In this study, we select only variables included in preprocessed data provided by the~authors.

\item \textit{metaMIMIC.} Dataset extracted from  the~MIMIC-IV database ~\citep{johnson2020mimic, johnson2023mimic} according to~\citep{woznica2022consolidated}. It contains a~collection of~12 binary classification tasks of~occurrence-specific diseases reported as ICD codes. The~MIMIC-IV database is the~most common resource of~high-volume EHR data.

\end{itemize}

In each case, we used preprocessed datasets containing subsets of~all available variables.

\end{document}